\documentclass[conference]{IEEEtran}
\IEEEoverridecommandlockouts

\usepackage{graphics} 
\usepackage{mathptmx} 
\usepackage{times} 
\usepackage{amsmath} 
\usepackage{amssymb}  
\usepackage{url}
\usepackage{graphicx}
\usepackage{float}
\usepackage{cite}
\usepackage{hyperref} 
\usepackage{cuted}
\usepackage{capt-of}
\usepackage{dsfont}

 \hypersetup{ 
     colorlinks=true, 
     linkcolor=black, 
     filecolor=black, 
     citecolor=blue,       
     urlcolor=black, 
     }

\title{\LARGE \bf
ForestSim: A Synthetic Benchmark for Intelligent Vehicle Perception in Unstructured Forest Environments
\vspace{-5pt}
}

\author{Pragat Wagle, Zheng Chen, Lantao Liu 
\thanks{All authors are with the Luddy School of Informatics, Computing,
and Engineering, Indiana University, Bloomington, IN 47408, USA. Email:
\{pwagle, zc11, lantao\}@iu.edu. \newline
The paper is accepted to 2026 IEEE Intelligent Vehicles Symposium (IV). 
}}

\begin{document}

\maketitle
\thispagestyle{empty}
\pagestyle{empty}

\begin{strip}\centering \vspace{-65pt}
{
\includegraphics[width=\linewidth]{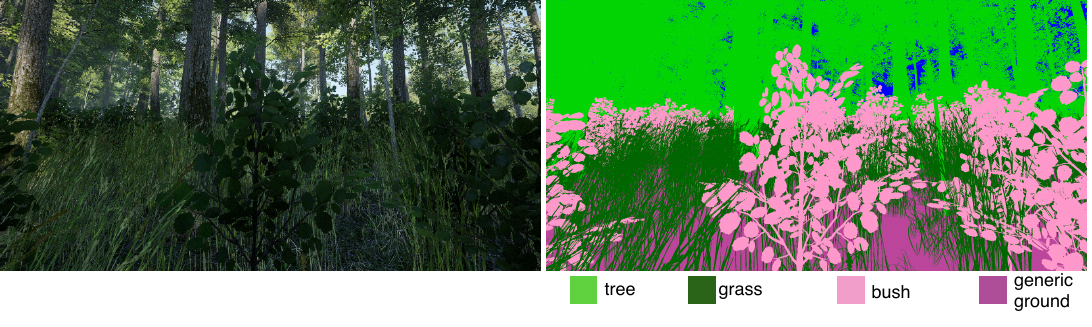} \vspace{-30pt}
\captionof{figure}{ An example image in \textbf{ForestSim} dataset. The size, shapes, and ratios are similar to those in the real world and demonstrate characteristics of an unstructured environment where the edges of objects are challenging to discern and almost blend into one another.
\vspace{-10pt}
\label{fig1:exam}}
}
\end{strip}

\begin{abstract}
Robust scene understanding is essential for intelligent vehicles operating in natural, unstructured environments. While semantic segmentation datasets for structured urban driving are abundant, the datasets for extremely unstructured wild environments remain scarce due to the difficulty and cost of generating pixel-accurate annotations. These limitations hinder the development of perception systems needed for intelligent ground vehicles tasked with forestry automation, agricultural robotics, disaster response, and all-terrain mobility.
To address this gap, we present ForestSim, a high-fidelity synthetic dataset designed for training and evaluating semantic segmentation models for intelligent vehicles in forested off-road and no-road environments. ForestSim contains 
2094 photorealistic images across 25 diverse environments, covering multiple seasons, terrain types, and foliage densities. 
Using Unreal Engine environments integrated with Microsoft AirSim, we generate consistent, pixel-accurate labels across 20 classes relevant to autonomous navigation. We benchmark ForestSim using state-of-the-art architectures and report strong performance despite the inherent challenges of unstructured scenes.
ForestSim provides a scalable and accessible foundation for perception research supporting the next generation of intelligent off-road vehicles. The dataset and code are publicly available:
\\ 
Dataset: \url{https://vailforestsim.github.io}\\ 
Code: \url{https://github.com/pragatwagle/ForestSim}
\end{abstract}

\section{Introduction}
\label{sec:intro}


Intelligent vehicles rely heavily on robust perception systems capable of interpreting diverse and dynamic environments. While significant progress has been made in urban autonomous driving supported by diverse, meticulously annotated datasets, achieving reliable autonomy beyond structured roads remains a major challenge. For example, many
wild environments, including forests, agricultural fields, and natural terrain, present complexities such as irregular geometry, dense vegetation, indistinct object boundaries, and seasonal variations. These factors create substantial difficulty for image segmentation pipelines that intelligent vehicles depend on for environment understanding, obstacle avoidance, and traversability estimation.

The advancement of computer vision systems is closely tied to the availability of large-scale image datasets. Such datasets have served as benchmarks and foundational resources for object detection, classification, and semantic segmentation, particularly when object classes are well represented in their environments \cite{ARCV,segbenchmark,burst,fisyscapes,semsegimprove1}.
The taxonomy of large datasets varies depending on their intrinsic characteristics. One popular taxonomy scheme categorizes image datasets into structured and unstructured datasets. Structured datasets, reminiscent of urban environments with clear boundaries, vehicular traffic, and regular-shaped buildings \cite{cocostuff, cityscapes,pascalvoc,adek,pascalcontext}, have been crucial in constructing high-performing semantic segmentation models. Furthermore, datasets are employed to augment existing models, enhancing the accuracy of semantic segmentation tasks. On the other hand, unstructured datasets exhibit significant variations in geometry, terrain, and appearance, presenting challenges such as navigational ambiguities in rough terrain and tall grass~\cite{charc1,RUGD2019IROS,tas500,drivelite,rellis}.

The creation of large pixelwise semantic labels poses substantial challenges, necessitating human intervention to ensure accuracy and quality. For instance, the CamVid dataset required 60 minutes per image for high-quality labeling \cite{BROSTOW200988}, while the Cityscapes dataset demanded 90 minutes per image~\cite{cityscapes}. Despite meticulous annotation efforts, generating pixel-accurate annotations often yields smaller datasets, particularly notable in high-quality semantic segmentation datasets. Beyond annotation challenges, navigating unstructured off-road environments presents additional hurdles, including resource scarcity and navigational complexities.

This paper explores the utilization of commercial environments built using Unreal Engine to generate pixel-accurate ground truth data for training semantic segmentation models. Leveraging the abundance of environments available within Unreal Engine, ranging in scale, coupled with the Microsoft-owned open-source tool AirSim, allows for the collection of unprocessed semantic segmentation images with persistent random labels across different instances. The ForestSim dataset, introduced herein, aims to enhance the accuracy of semantic segmentation models across various terrain types, leveraging diverse environments representative of different seasons. The dataset comprises 2094 RGB images with corresponding pixel-wise ground truth annotations, extracted from 25 different high-quality, realistic environments. Although ForestSim is smaller in scale its focus on unstructured forested seasonal environments with pixel accurate annotation provides a complimentary resource for studying off-road perception challenges. An example RGB image and ground truth pixel-level semantic segmentation image produced from the data collected from Unreal Engine are illustrated in Fig.~\ref{fig1:exam}. Benchmarks, employing up-to-date methodologies such as Mean Intersection-Over-Union and Pixel Accuracy metrics \cite{RUGD2019IROS, rellis}, validate the efficacy of the proposed dataset.

We introduce this dataset providing pixel-accurate semantic labels focusing exclusively on densely vegetated forested, seasonal, and off-road scenes and establish baseline performance with the goal of enhancing the capabilities of existing machine learning models. By leveraging this dataset, we see potential to support autonomous field robots in a myriad of tasks, including timber sorting, harvesting operations, agricultural field tasks, and surveillance in challenging, unstructured environments. We leave environment-disjoint and cross-environment evaluation as future work.  

\section{Segmentation Datasets}
\label{sec:existing}

Semantic segmentation datasets serve as foundational resources for partitioning images into meaningful parts through pixel-wise annotation. This segmentation task is fundamental for various applications in computer vision, facilitating tasks such as object detection, scene understanding, and autonomous navigation \cite{Feng_2021}. These pipelines include utilizing an encoder to create hierarchical representations of an image using a backbone such as ResNet with a decoder that upsamples samples through convolutional layers, converting low dimensional features to original resolutions, which are the feature maps that are eventually used for pixel-wise prediction.

\subsection{Structured Datasets}
Structured semantic segmentation datasets represent environments with well-defined boundaries and organized elements. This category encompasses a plethora of datasets, including COCO-Stuff \cite{cocostuff}, Pascal VOC \cite{pascalvoc}, ADE20K \cite{adek}, Pascal Context \cite{pascalcontext}, Audi Structured \cite{audiStuctured}, Cityscapes \cite{cityscapes}, KITTI \cite{KITTI}, Mapillary \cite{ertler2020mapillary}, and ApolloScape \cite{apolloscape}.
Among numerous selections, for example, Mapillary provides a benchmark dataset specifically tailored for traffic sign classification \cite{ertler2020mapillary}, while KITTI focuses on common urban objects like buildings, trees, cars, and roads \cite{KITTI}. ApolloScape offers a diverse range of data captured from various cities and different times of the day, integrating camera videos, consumer-grade motion sensors, and 3D semantic maps \cite{apolloscape}. These datasets are often collected by involving ground vehicles equipped with multiple sensors, capturing rich data from urban environments \cite{cityscapes, apolloscape, KITTI}.

\subsection{Unstructured Datasets}
In contrast, unstructured semantic segmentation datasets capture environments with complex and varied characteristics, including rugged terrain, dense vegetation, and irregular structures. The RUGD dataset \cite{RUGD2019IROS} serves as a benchmark for unstructured environments near creeks, vegetation, water bodies, trails, and villages. TAS500 \cite{tas500} focuses on discerning traversable regions from non-traversable ones, categorizing 44 different objects into nine groups. The Rellis dataset \cite{rellis} comprises synchronized sensor data collected using a mobile robotic platform, featuring diverse terrains like runways, aprons, and lakes. These datasets play a crucial role in advancing the robustness and adaptability of semantic segmentation models in challenging real-world scenarios.

However, compared to the structured datasets, the number of unstructured datasets is significantly lower. This scarcity stresses the importance of creating more datasets in this category to provide a more comprehensive representation of diverse and challenging environments. In our work, we provide a very accessible and reusable process to improve upon this scarcity through the use of simulated environments.

\section{Relevant Uses in Autonomy}
\label{sec:relevant_uses}

Understanding the characteristics of an environment is instrumental in various autonomous applications, particularly in supporting robot path estimation and navigation tasks. Leveraging both 3D terrain information and visual features collectively yields superior results compared to relying on either resource alone \cite{uses1}. Models can be devised to generate color images and assign traversability costs to different regions based on their geometric attributes and visual appearance, contributing to more informed decision-making processes \cite{uses2}. Texture-based features derived from onboard sensors such as IMU, motor current, and bumper switches aid in binary segmentation of terrain traversability, enhancing the vehicle's ability to navigate challenging terrains \cite{uses3}. Additionally, learning approaches that utilize models trained on data collected at different time points can improve nearsightedness by referencing past trajectory data \cite{uses3}.

Domain adaptation offers a promising pathway for integrating ForestSim with real-world datasets by mitigating domain discrepancies in semantic segmentation. Unsupervised Domain Adaptation (UDA) leverages labeled source data and unlabeled target data to reduce domain gaps through feature alignment, input-level adaptation, image transfer, and discriminator-based learning \cite{da1,da2,da3,da4,da5,da6,da8,da9}. By simplifying data preparation, these techniques enhance the robustness and adaptability of segmentation models in complex environments \cite{lee2020unsupervised}.

\section{ForestSim Data Collection and Preparation}
\label{sec:data}

Synthetically generated data has emerged as a powerful tool for enhancing the performance of deep neural networks in image segmentation tasks. Benchmark evaluations conducted on synthetic datasets have demonstrated comparable accuracy to real-world data in image segmentation tasks. Moreover, with the application of domain adaptation techniques, synthetic data can not only mimic but also outperform real-world datasets, thereby broadening the scope of dataset applications \cite{shafaei2016play}.

\begin{figure}[t]
{
\centering
\includegraphics[width=8.5cm, height=11.5cm]{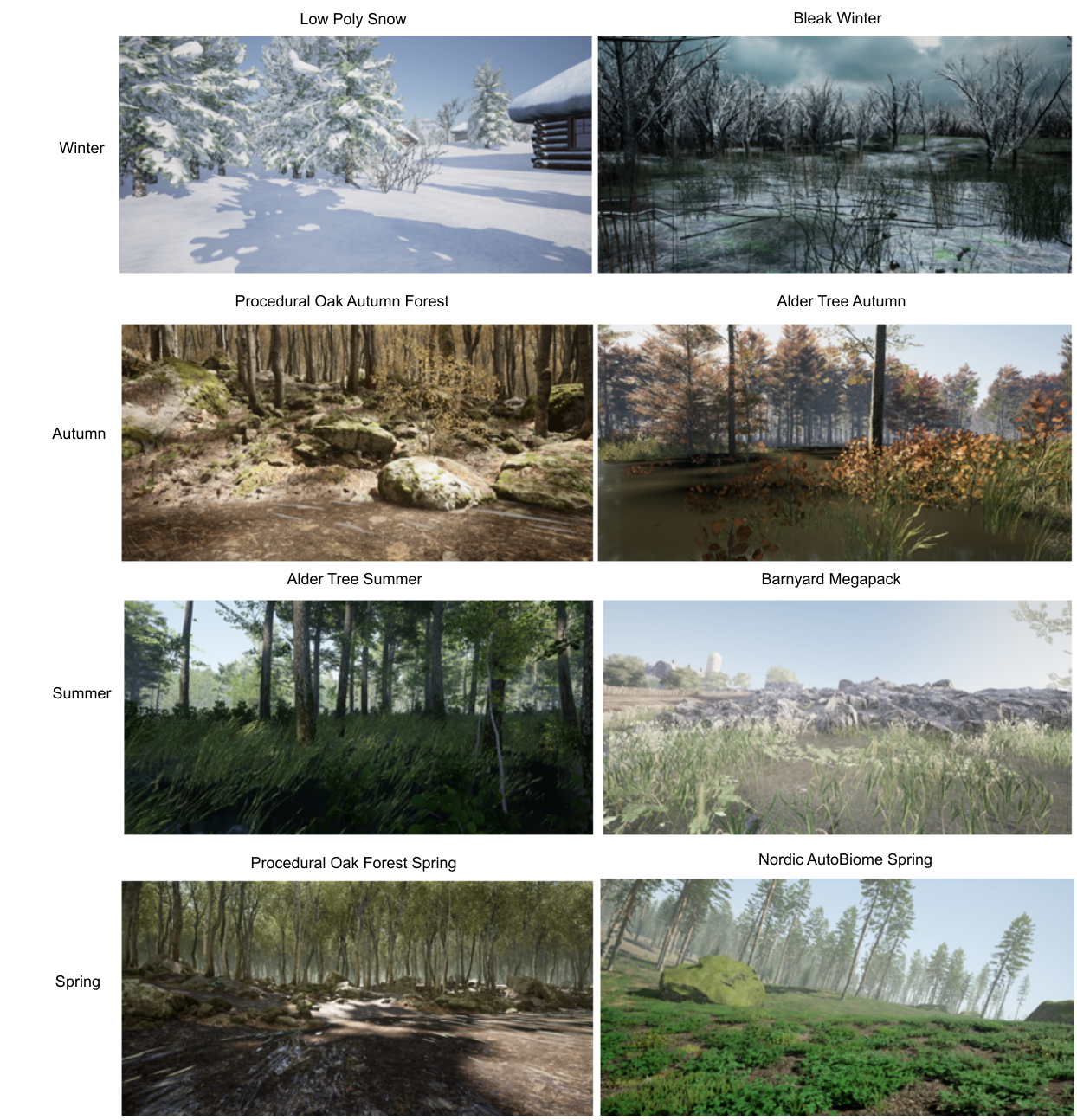} \vspace{-10pt}
\caption{Example RGB images of seasonal environments. These pictures demonstrate the unstructured, off-road, and forested characteristics of complex environments.\vspace{-10pt}}
\label{fig2:rgb}
}
\end{figure}

\subsection{ForestSim Environments}
The proposed ForestSim dataset includes a diverse array of environments, ranging from mountains, forests, and hills to jungles and marshes. As depicted in Fig.~\ref{fig2:rgb}, example images showcase the varied terrain types present in the dataset. 
This diversity extends to the geometry and size of objects within the environments, facilitating the development of more adaptable models compared to datasets with limited diversity. Notably, the environments utilized in our ForestSim are meticulously crafted for commercial purposes, with high-fidelity realism in appearance, proportions, lighting, textures, and object placement. 
The unreal engine provides photorealistic environments with various illuminations and changing light conditions.
This level of visual fidelity enhances the dataset’s utility for studying synthetic-to-real transfer in off-road perception tasks \cite{qiu2016unrealcv}.

\subsection{Hardware and Software}
Data collection involves a combination of manual intervention and automation, presenting unique challenges. Similar to the TartanAir dataset \cite{tartanair}, our approach integrated various modalities, including RGB images and segmentation data.

The data collection system operates on hardware comprising an Intel NUC NUC11PHKi7 11th Gen Core i7-1165G7 Quad-Core processor, 32GB DDR4 RAM, 1TB PCIe NVMe SSD, and GeForce RTX 2060 6GB GDDR6 Graphics, running the Windows 11 OS. Both Unreal Engine and AirSim \cite{AirSim2017fsr} offer robust support for Windows and macOS environments. Notably, hardware limitations can impact performance, emphasizing the importance of optimizing system configurations.

The Epic Games Launcher is leveraged to install Unreal Engine and access environments, while simulation within Unreal Engine is facilitated by AirSim, a powerful plugin. AirSim enables interaction with ground or air vehicles programmatically, offering functionalities such as image retrieval, state querying, and vehicle control. Interactions with the AirSim API are orchestrated using Python 3.7, ensuring seamless integration and flexibility in data collection processes.

\begin{figure}[t] \vspace{-2.5pt}
{
\centering
\includegraphics[width=8.0cm, height=3.0cm]{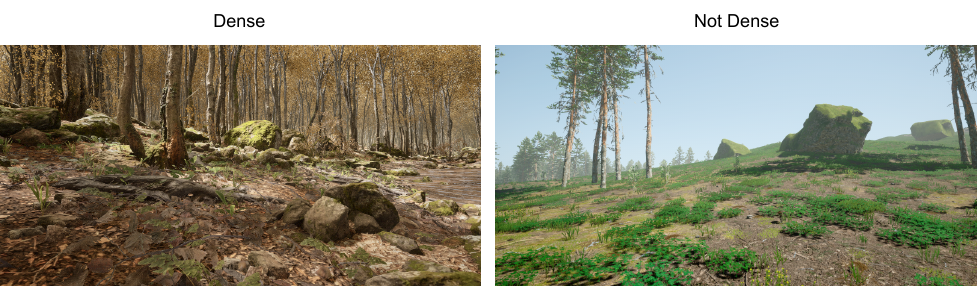}
\vspace{-10pt}
\caption{More dense environments, an example is seen on the left, required manual control. On the right, data was able to be collected programmatically with no manual control. \vspace{-10pt}}
\label{fig3:dense}
}
\end{figure}

\subsection{Data Acquisition}
During data collection manual intervention was required only in a small subset of cases, primarily during navigation in denser regions, while the majority of images were generated fully automatically. At five-second intervals, RGB and segmentation images are captured by a ground vehicle outfitted with three cameras—front left, front center, and front right—using AirSim.

The pixel RGB assignment propagated through environments, after initial processing, substantially decreasing labeling time compared to CamVid and Cityscapes. 

The vehicle follows predefined paths, synchronized with time intervals, optimizing efficiency in its given operating environments. However, navigation poses challenges in congested areas where small, impassable objects increase collision risks, occasionally necessitating manual control to resolve navigation issues. Fig.~\ref{fig3:dense} illustrates a scenario of such challenges encountered during data acquisition.

\subsection{Data Processing and Statistics}
Data processing primarily centers on segmentation images collected via AirSim, aiming at developing and annotating pixel-wise ground truth labels. AirSim assigns a unique ID to each static mesh, mapping it to an RGB value from a predefined palette of 255 RGB values. ForestSim scenes are static, and occlusions or overlaps are resolved by the Unreal Engine rendering pipeline based on the foremost visible surface at each pixel. However, inconsistencies arose in object labeling and color assignments across different environments. 
To address this, each environment underwent manual curation, establishing mappings to reconcile variations in object labeling and RGB assignments. For instance, disparate RGB values assigned to the same object class are consolidated, ensuring uniformity across environments. Fig.~\ref{fig6:ground} presents the finalized mappings of object classes to RGB values, with trees serving as an illustrative example.


\begin{figure}[t]
{
\centering
\includegraphics[width=8.5cm, height=11cm]{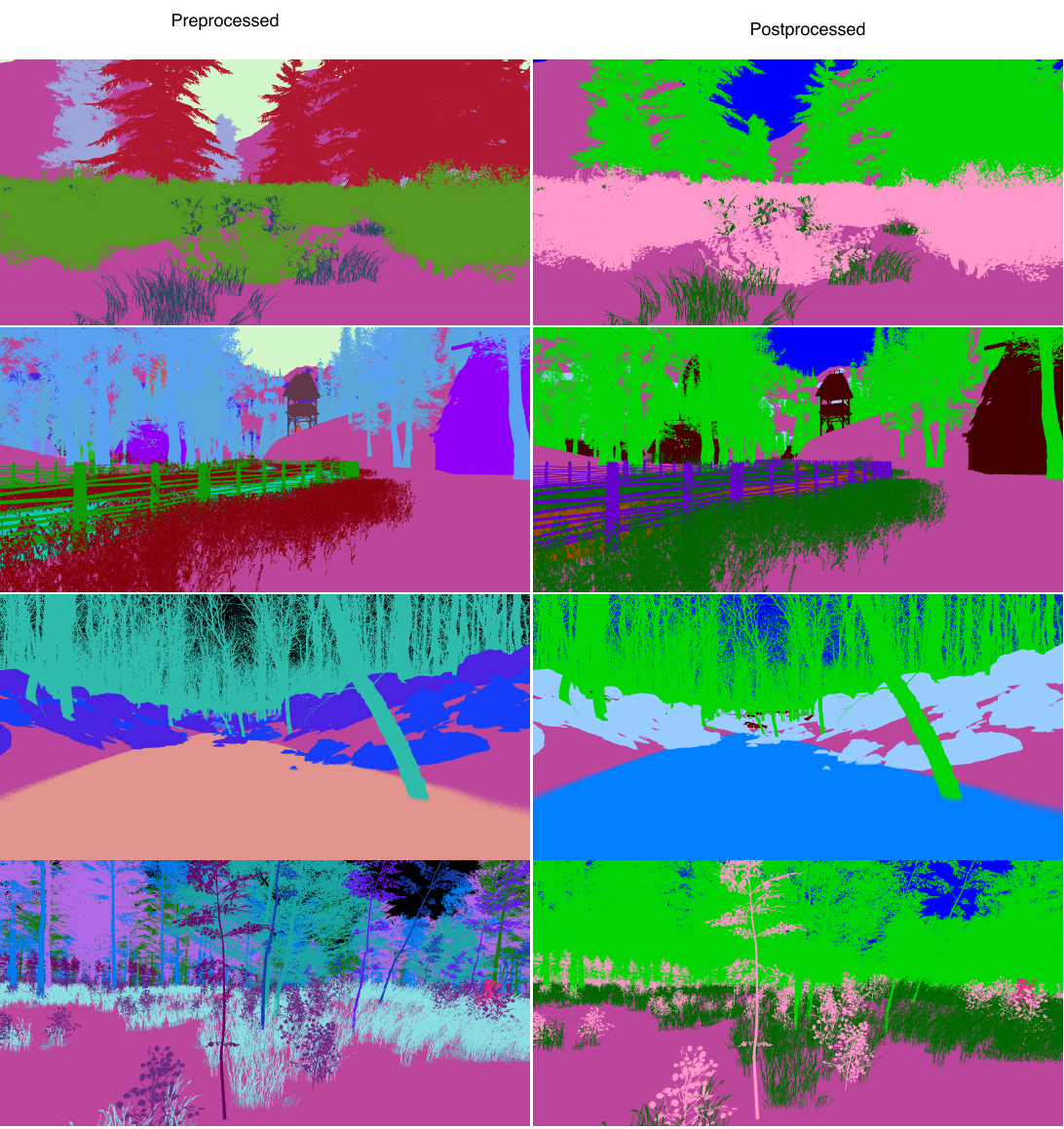} \vspace{-5pt}
\caption{Examples of segmentation images captured directly from AirSim are on the left. These images were processed by manually determining the object each RGB value corresponds with and using this mapping to generate the ground truth pixelwise labels on the right. \vspace{-10pt}}
\label{fig4:examseg}
}
\end{figure}

The semantic labels for our dataset are established through meticulous mapping and reconciliation, eliminating redundancy and harmonizing object representations across diverse environments. Fig.~\ref{fig4:examseg} showcases examples of original and converted segmentation images, highlighting the efficacy of our data processing pipeline. Our pipeline required mapping for each unique environment. The time-intensive process was creating the mapping, which required manually examining all of the uniquely colored pixels within the collected segmentation images. After mapping was complete, the consolidation process of relabeling the individual pixels was automated. 

\begin{figure}[t]
{\centering
\includegraphics[width=8.5cm, height=6.5cm]{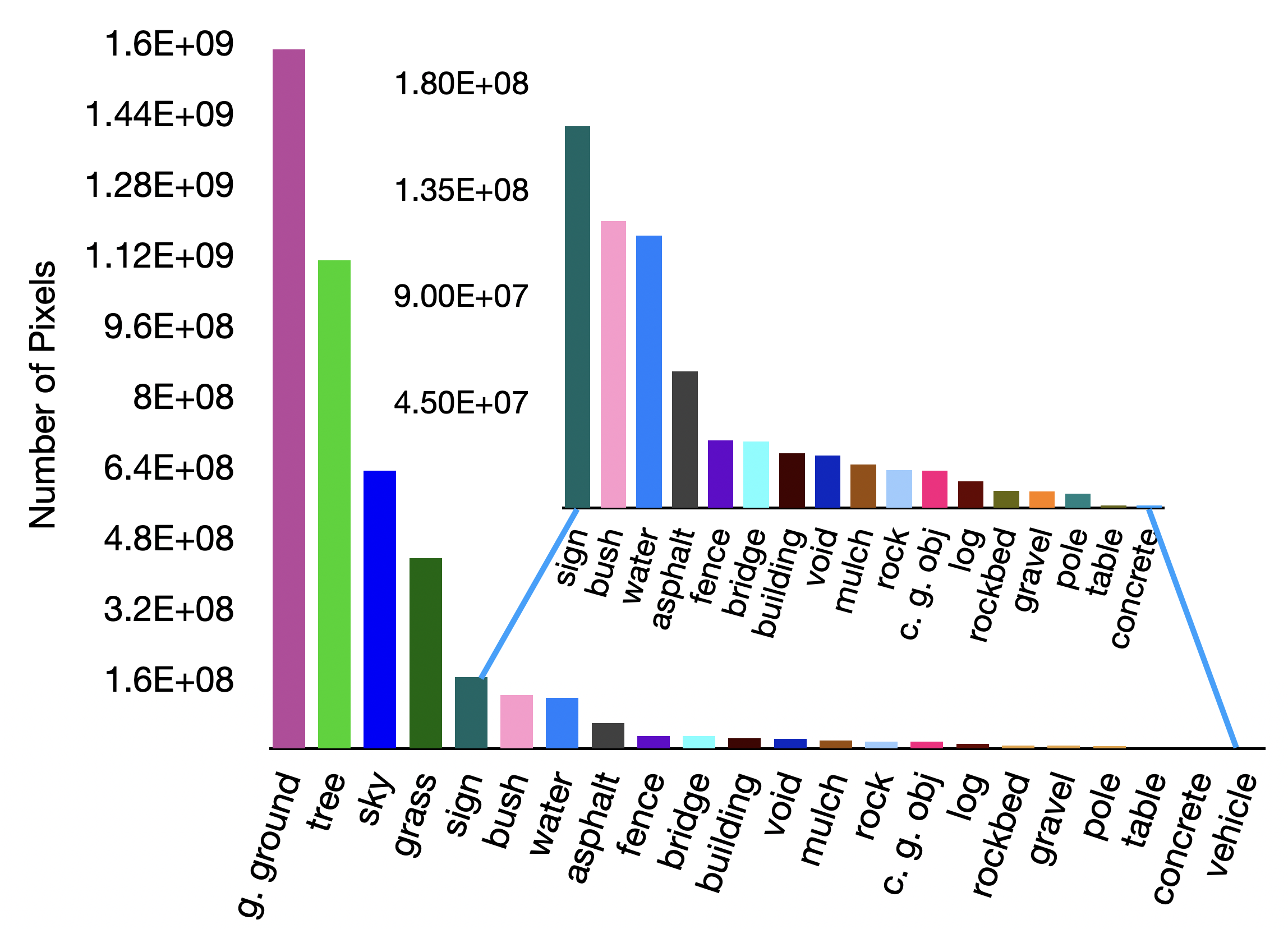} \vspace{-10pt}
\caption{Numbers of total pixels per class in the dataset in descending order. \vspace{-15pt}}
\label{fig5:total}
}
\end{figure}

\section{Annotation Statistics and Ontology}

Our ForestSim dataset consists of a diverse array of classes essential for semantic segmentation, including grass, trees, poles, water bodies, sky, vehicles, containers, asphalt, gravel, mulch, rock beds, logs, bushes, signs, rocks, bridges, concrete structures, buildings, void regions, and generic ground. Fig.~\ref{fig5:total} provides an overview of the distribution of pixels across these classes within the dataset.

Each of the 20 distinct object classes is assigned a unique RGB value for identification and labeling. The category of ``generic ground" comprises all traversable ground surfaces, which may obscure fine grained distinctions but reflects practical navigation-oriented semantics. In AirSim, flat ground in certain environments was labeled a specific RGB value, most likely because no static mesh was used for it during development. These are traversable, flat regions. 
Moreover, the category of ``generic container objects" includes a variety of miscellaneous objects that may pose collision risks or influence navigation. These include benches, trash cans, playground equipment (such as slides and swings), water containers, log containers, and similar items.

 \begin{figure*} [t] 
 {
    \centering
    \includegraphics[width=\textwidth]{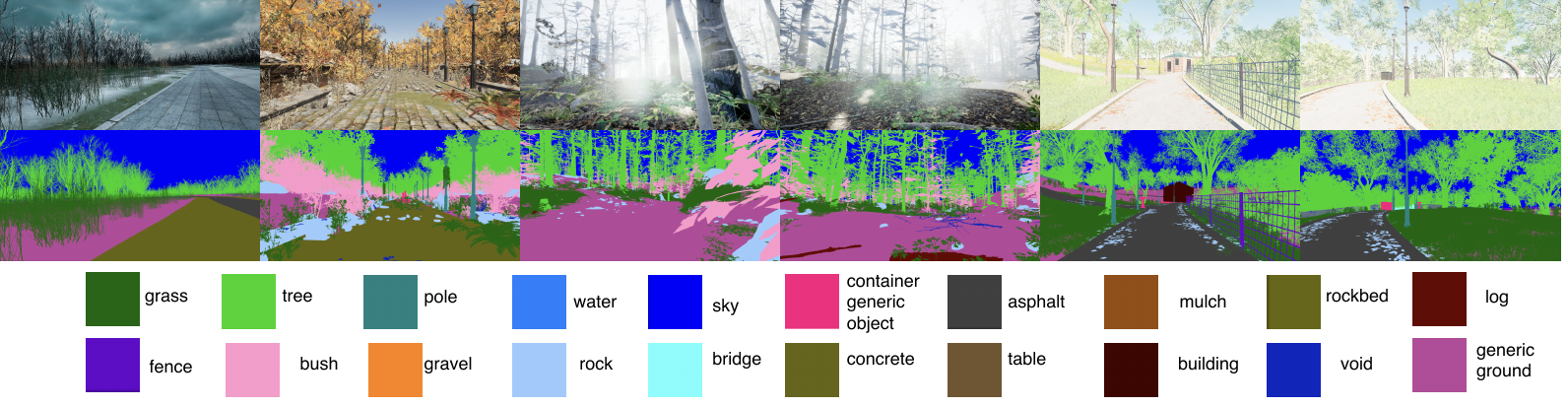}\vspace{-10pt}
\caption{Examples of ground truth annotations from the ForestSim dataset. The first row is the photorealistic RGB image collected from the environment, and the second row is the corresponding semantic segmentation. Please note that these are the pixel-wise, true semantic segmentation images after consolidation and labeling. \vspace{-10pt}}
    \label{fig6:ground}
}
\end{figure*}

\begin{figure*}[t] 
{
    \centering 
    \includegraphics[width=\textwidth]{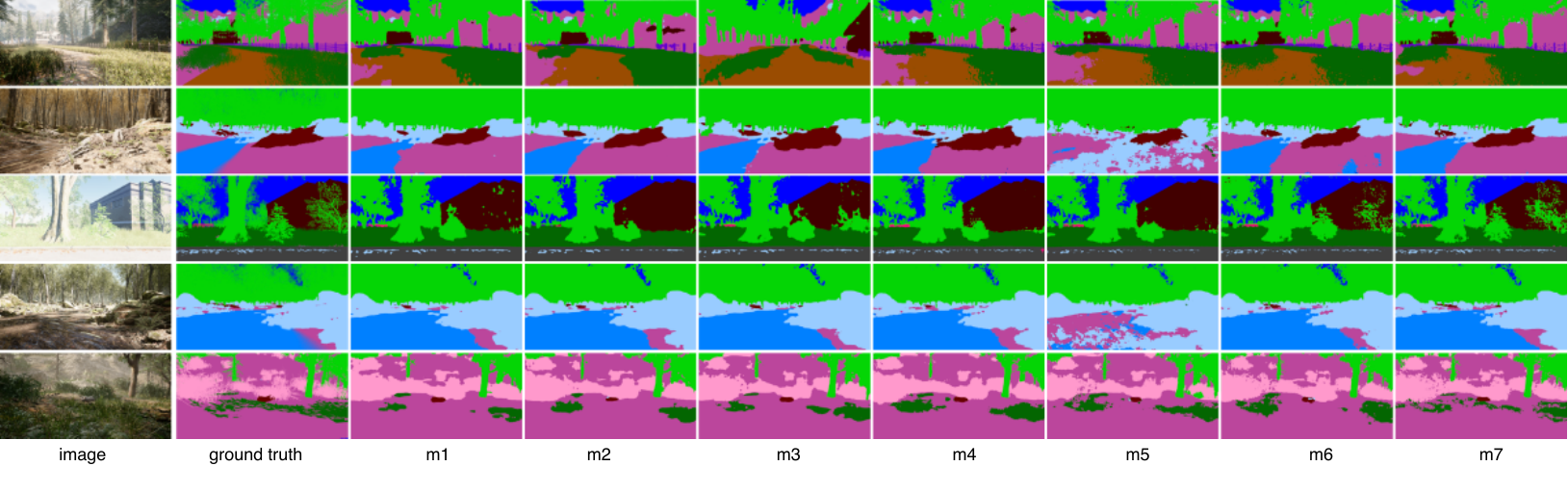}\vspace{-10pt}
    \caption{The original image, the ground truth, and the predicted image annotation for models 1 to 7.  \vspace{-10pt}}
    \label{fig:model1-7}
}
\end{figure*}

\begin{figure*}[!htb]
{
    \centering 
    \includegraphics[width=\textwidth]{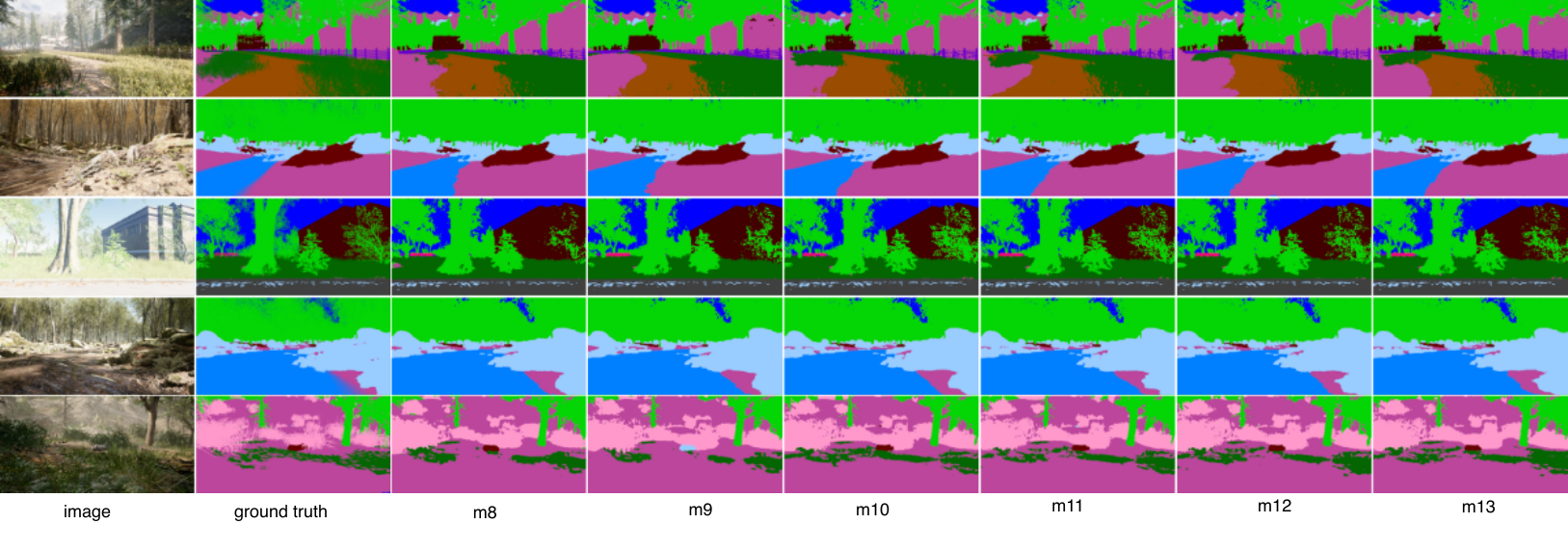}\vspace{-10pt}
        \caption{The original image, the ground truth, and the predicted image annotation for models 8 to 13. \vspace{-25pt}}
    \label{fig:model8}
}
\end{figure*}

Despite the comprehensive coverage of classes, data sparsity is observed in certain categories, such as vehicles, concrete structures, poles, gravel, and rock beds, constituting a minimal percentage of the dataset. This sparsity presents challenges in accurately classifying these objects, potentially leading to erroneous decisions during segmentation tasks. 
Additionally, dynamic scenarios are not readily simulated within AirSim, limiting the availability of data capturing such situations. However, existing methodologies can address these limitations, offering avenues for enhancing dataset diversity and mitigating segmentation challenges in ForestSim.

\section{Benchmarks for Domain Adaptive Segmentation}
\label{benchmarks}

\subsection{Baselines and Experimental Setups}
The benchmarking process for ForestSim leverages a unified framework implemented using mmsegmentation \cite{mmseg2020} and focuses on evaluation representative modern segmentation architectures on ForestSim to establish baseline performance. Models are structured following an encoder-decoder pattern, with various configurations explored to optimize segmentation performance.

\begin{table*}[t] \vspace{-5pt}
\centering
\caption{Prediction results from architectures used beginning with the model number, the pretrained model, encoder, and decoder.}
\small
\label{tab:result}
\begin{tabular}{l|l|c|c|c}
\hline
 Model & Method & IOU$\downarrow$& Pix. Acc.$\downarrow$ & M. Pix. Acc.~$\downarrow$  \\
\hline
m1 & resnet50v1c + ResNetV1c + PSPHead & 61.64 & 89.85 & 72.14 \\
\hline
m2 & resnet50v1c + ResNetV1c + ASPPHead & 61.87 & 89.91 & 72.81 \\
m3 & resnet101v1c + ResNetV1c + ASPPHead & 62.81 & 89.86 & 73.13 \\
\hline
m4 & resnet50v1c + ResNetV1c + DepthwiseSeparableASPPHead & 59.16 & 89.31 & 72.93 \\
m5 & resnet101-v1c + ResNetV1c + DepthwiseSeparableASPPHead & 59.22 & 88.32 & 69.56 \\
\hline
m6 & mit-b0 + MixVisionTransformer + SegformerHead & 61.82 & 90.52 & 71.12 \\
m7 & mit-b5 + MixVisionTransformer + SegformerHead & 67.93 & 92.05 & 76.42 \\
\hline
m8 & resnet50 + ResNet + Mask2FormerHead & 67.48 & 91.34 & 75.77 \\
m9 & resnet101 + ResNet + Mask2FormerHead & 65.80 & 91.29 & 74.61 \\
m10 & swin-base + SwinTransformer + Mask2FormerHead & 74.50 & 92.57 & 82.30 \\
m11 & swin-large + SwinTransformer + Mask2FormerHead & 75.31 & 92.65 & 82.68 \\
m12 & swin-tiny + SwinTransformer + Mask2FormerHead & 70.46 & 92.14 & 79.79 \\
m13 & swin-small + SwinTransformer + Mask2FormerHead & 74.02 & 92.39 & 81.39 \\
\hline 
\end{tabular} 
\end{table*}

One approach utilizes a pretrained ResNet50v1c model as the encoder, coupled with a PSPNet decoder, employing Cross Entropy Loss with a weight of 1.0. Additionally, other models combine pretrained ResNet50v1c and ResNet101v1c models with different decoders, including Atrous Spatial Pyramid Pooling (ASPP) and DepthwiseSeparable, each utilizing Cross Entropy Loss with specific weight configurations. 
Furthermore, two models integrate MixVisionTransformer and Segformer decoders with Cross Entropy Loss, based on pretrained mit-b0 and mit-b5 models, respectively. Another set of models employ ResNet encoders paired with Mask2Former decoders, augmented with MSDeformAttnPixel and trained with various loss functions and optimizer settings. 
Additionally, models incorporating SwinTransformer decoders, built from combinations of pretrained Swin Tiny, Swin Small, Swin Base, and Swin Large models, are utilized. These models are trained using AdamW optimizer and PolyLR scheduler.

\subsection{Data Split, Training, and Evaluation Metrics}
The dataset was split randomly into training (90\%) and testing (10\%) sets to evaluate within-dataset generalization. Training of models occurred on 4 nodes, each containing SUSE Enterprise Linux Server (SLES) version 15 with 256 GB of memory and two 64-core, 2.25 GHz, 225-watt AMD EPYC 7742 processors running 4 tasks per node and 4 NVIDIA A100 GPUs per node. The number of iterations for training varies based on the scheduler that was used when configuring the models, but it ranged from 40,000 to 160,000 iterations.

We report Mean IoU and pixel accuracy as they are widely adopted baseline metrics in semantic segmentation. More fine-grained evaluations, such as per-class IoU and boundary-aware metrics, are valuable directions for future analysis, particularly for thin structures and class boundary ambiguity in unstructured environments. Mean IoU is the average IoU between all classes~\cite{metrics1}. The IoU for each class is computed as $\frac{TP}{TP+FP+FN}$. Mean pixel-wise segmentation accuracy is also used, which is the average segmentation accuracy per model and is a preferred metric as it evenly weights each class.

\subsection{Analysis and Experimental Evaluation}

The trained models were employed to make predictions on the randomized test set, and their performances were evaluated and summarized in Table~\ref{tab:result}. Notably, the ForestSim dataset stands out for its exceptional quality, with significant effort dedicated to preparing accurate ground truth labels. Rigorous review and refinement processes were implemented, ensuring the removal of low-quality data and enhancing the dataset's integrity as a high-quality baseline for training on an unstructured simulated environment. Performance variability across classes is partly attributable to class frequency imbalance, with smaller or visually ambiguous classes exhibiting reduced segmentation accuracy. Visual representations of prediction results from the trained models are illustrated in Fig.~\ref{fig:model1-7} and \ref{fig:model8}.

Table~\ref{tab:result} summarizes all of the model results on the test set which was a random 10\% of our data. All of these models were trained and tested on the same data. The table breaks down the methods that are built using a pretrained model, an encoder, and a decoder. Our results show the IoU ranging from a low of 59.16\% to 74.02\%. The IoU value ranges demonstrate that the model is performing relatively well and is predicting the majority of the class correctly. The unclear edges and boundaries of objects are also negatively impacting this result, which is one of the existing challenges of unstructured environments. Pixel accuracy, which is the total correct predicted divided by the total number of pixels, ranges from a low of 88.32\% to 92.65\%. We conclude that objects that are represented more in the data are being predicted with high accuracy. The mean pixel accuracy, which is the average prediction accuracy of all of the classes, was negatively impacted due to a 0 percent accuracy for vehicle. The mean pixel accuracy for concrete and table was the next two lowest after vehicle. That correlates with our conclusion that this is due to data sparsity, as these were the three least represented classes. Pole was able to be predicted well, but most likely due to its distinctive geometric structure. 

\subsection{Discussion and Future Work}

The findings are promising. Our future work will be further improvement through data balancing and enrichment of sparse classes. For instance, augmenting ForestSim with complementary datasets shows promise in enhancing the adaptability of semantic segmentation models. Integration with diverse simulation environments or existing datasets can address challenges such as dynamic behavior and data sparsity. Moreover, leveraging synthetic image production and GAN networks for domain transfer between synthetic and real-world datasets holds considerable potential for gaining valuable insights and making significant improvements.
\section{Conclusion}
We introduce ForestSim, a synthetic semantic segmentation dataset specifically designed for intelligent vehicle perception in unstructured  environments. ForestSim offers realistic seasonal variation, diverse terrain, and consistent pixel-wise labels across 20 classes critical for autonomous navigation. This dataset comprises 2094 ground truth pixel-wise annotations, providing a valuable resource with high accuracy for semantic segmentation tasks. 
Extensive experiments show that ForestSim provides a robust baseline for training state-of-the-art segmentation models and is well-suited for advancing intelligent vehicle research beyond structured on-road environments.

\bibliographystyle{unsrt}
\bibliography{ref}

\end{document}